# InvNeRF-Seg: Fine-Tuning a Pre-Trained NeRF for 3D Object Segmentation

Jiangsan Zhao[1], Jakob Geipel[1], Krzysztof Kusnierek[1], Xuean Cui[2*]

*Abstract*—Neural Radiance Fields (NeRF) have been widely adopted for reconstructing high-quality 3D point clouds from 2D RGB images. However, the segmentation of these reconstructed 3D scenes is more essential for downstream tasks such as object counting, size estimation, and scene understanding. While segmentation on raw 3D point clouds using deep learning requires labor-intensive and time-consuming manual annotation, directly training NeRF on binary masks also fails due to the absence of color and shading cues essential for geometry learning. Existing NeRF-based 3D segmentation methods fall into two categories: SA3D and FruitNeRF. In this work, we implemented a field-density-based variant of SA3D to generate denser 3D segmentations. While SA3D remains a fast post-processing approach that back-projects 2D masks into the 3D space, its outputs are often noisy due to lack of density field optimization. FruitNeRF introduces a segmentation head and binary mask supervision during training, but the joint learning process can hinder both RGB and segmentation performances. To address these limitations, we propose Invariant NeRF for Segmentation (InvNeRF-Seg), a two-step, zero-change fine-tuning strategy for 3D segmentation. We first train a standard NeRF on RGB images and then fine-tune it using 2D segmentation masks—without altering either the model architecture or loss function. This approach produces higher-quality, cleaner segmented point clouds directly from the refined radiance field with minimal computational overhead or complexity. Field density analysis reveals consistent semantic refinement: densities of object regions increase while background densities are suppressed, ensuring clean and interpretable segmentations. We demonstrate InvNeRF-Seg's superior performance over both SA3D and FruitNeRF on both synthetic fruit and real-world soybean datasets. This approach effectively extends 2D segmentation to high quality 3D segmentation. Our code will be available at https://github.com/ZJiangsan/InvNeRF-Seg.

*Index Terms*—Neural Radiance Fields, fine tuning, 3D segmentation

## I. Introduction

NEURAL Radiance Fields (NeRF) have demonstrated remarkable success in reconstructing photo-realistic 3D scenes from 2D images by learning a volumetric scene representation [1], [2]. However, while NeRF-based models have been widely adopted for view synthesis, research into object-level 3D segmentation within reconstructed NeRF scenes remains relatively limited [3], [4], [5].

Accurate 3D segmentation is essential for downstream applications such as robotic perception [6], augmented reality (AR)[7], autonomous driving[8], and medical imaging[9] and plant phenotyping in agriculture [5], [10]. However, existing approaches to 3D segmentation either rely on fully annotated point clouds—requiring substantial manual labeling [11]—or introduce significant architectural complexity by coupling additional semantic branches or loss functions into NeRF models [4], [5]. Alternatively, post-processing-based methods like SA3D [3] perform 3D segmentation by back-projecting 2D masks, but suffer from noise and lack of density refinement. These limitations hinder the adoption of NeRF-based segmentation in real-world applications where robustness and ease of implementation are critical.

Motivated by these challenges, we introduce InvNeRF-Seg, a simple yet effective two-step fine-tuning method for NeRF-based 3D object segmentation. As the name suggests, the method maintains the original NeRF architecture and loss function entirely unchanged. The first step incolves training a high quality NeRF model on multi-view RGB images using a standard NeRF pipeline. In the second step, the pretrained NeRF is fine-tuned using 2D segmentation masks of the same images—without altering any architectural components—to derive a 3D segmentation-aware model.

Despite its simplicity, InvNeRF-Seg produces high-quality segmented 3D point clouds across both synthetic fruit datasets and real-world plant imagery, offering a practical bridge between 2D mask supervison and 3D semantic reconstruction.

The main contributions of this work are summarized as follows:

- We introduce InvNeRF-Seg, a two-stage fine-tuning strategy that enables 3D object segmentation from a pretrained NeRF model without modifying its architecture or loss function.
- We demonstrate that InvNeRF-Seg produces accurate, high-quality segmented 3D point clouds, enabling object-level tasks such as counting and localization.

This study was partially supported by Biological Breeding-National Science and Technology Major Project (2023ZD04076), China, the Norwegian Agricultural Agreement Research Fund/Foundation for Research Levy on Agricultural Products in the framework of RASK project [grant number 352849], Grofondet [grant number 240069], and the Research Council of Norway base funding [contract No. 342631/L10].

Authors [1] are with Department of Agricultural Technology, Center for Precision Agriculture, Norwegian Institute of Bioeconomy Research (NIBIO), Nylinna 226, 2849, Kapp, Norway and [2] is from Biotechnology Research Institute, Chinese Academy of Agricultural Sciences, Beijing 100081, China (*corresponding author). E-mails: jiangsan.zhao@nibio.no; jakob.geipel@nibio.no; krzysztof.kusnierek@nibio.no; cuixuean@caas.cn.

- We conduct comprehensive evaluations comparing InvNeRF-Seg to SA3D and FruitNeRF using standard metrics including Intersection over Union (IoU) and Peak Signal-to-Noise Ratio (PSNR).
- We analyze the radiance field before and after fine-tuning, showing consistent semantic refinement where object densities are reinforced and background is suppressed.
- We further show that InvNeRF-Seg naturally extends to multi-class segmentation without requiring any changes to model structure or training objectives.

The rest of the paper is organized as follows: Section II reviews related work on NeRF-based 3D segmentation. Section III presents the details of the proposed InvNeRF-Seg method. Section IV provides a comprehensive evaluation of the approach. Section V discusses key insights and practical considerations. Finally, conclusions are drawn in Section VI.

## II. RELATED WORK

Classical deep learning methods for 3D segmentation, such as PointNet [11], PointNet++[12], have achieved remarkable success. However, their reliance on labor-intensive and time-consuming 3D point cloud annotation remains a significant bottleneck, limiting their scalability and applicability across domains [13]. Additionally, these methods often struggle in scenarios with noisy or incomplete point clouds, especially in complex environments with occlusions [14], [15].

Neural Radiance Fields (NeRF) have transformed 3D reconstruction through learning implicit neural representations of scenes from multi-view RGB images, enabling photo-realistic novel view synthesis [2]. Building on this foundation, several approaches have been developed on top of NeRF to achieve 3D segmentation with the help of segmented 2D segmentation masks. Among these, two prominent categories have emerged: Segment Anything in 3D (SA3D) [3] and FruitNeRF [5].

SA3D is a post processing method that does not change the NeRF model or training. Instead, it uses a pretrained RGB NeRF and performs 3D segmentation by projecting 2D binary masks into the scene using the rendered weights $\omega_i$ for each sample point $X_i$ along ray $r$.

The weights from volume rendering are computed as:
$$\omega_i = T_i \cdot (1 - \exp(-\sigma_i \delta_i)) \cdot m_i$$
where $T_i$ is the accumulated transmittance, $\sigma_i$ is the predicted density, $\delta_i$ is the distance between sampled points, and $m_i$ is the binary mask value for the corresponding 2D pixel. For all rays $r$ whose corresponding 2D pixels are labeled as foreground in the segmentation mask (i.e. $M(r) = 1$, SA3D collects 3D points where the rendering weights exceed a threshold $\tau$:
$$P_{SA3D} = \{X_i | \sum_r M(r) \cdot \omega_i(r) > \tau\}$$
No backpropagation or model update is involved in SA3D.

Even though it is fast, the noisy and fragmented segmentation point cloud affects subsequent analysis negatively because it is harshly encoded, and no additional learning is involved for its field density optimization. Although SA3D incorporates self-prompting strategies and IoU-aware view rejection to improve segmentation robustness, these techniques primarily address errors in 2D mask generation. High quality 2D mask generation is a preprocessing step for high quality NeRF segmentation model and shouldn't interfere the capability of an effective NeRF segmentation model.

FruitNeRF, on the contrary, is a joint training of NeRF for RGB and segmentation of fruits. FruitNeRF extends NeRF by adding a segmentation branch that outputs a scalar foreground probability $m(t) \in [0,1]$ at each sampled point along a ray. The output of the modified NeRF is:
$$f_\theta(X, d) \rightarrow (c, \sigma, m)$$
c: RGB color, $\sigma$: volume density, m: segmentation confidence. The rendered mask is computed with the same volume rendering accumulation used for RGB color:
$$\widehat{M}(r) = \sum_{i=1}^{N} T_i(1 - \exp(-\sigma_i \delta_i)) \cdot m_i$$
Then FruitNeRF uses a composite loss:
$$l_{FruitNeRF} = l_{RGB} + \varphi l_{seg}$$
where $l_{RGB}$: mean squared error (MSE) loss on RGB reconstruction; $l_{seg}$= BCE($\widehat{M}$, $M_{gt}$): binary cross-entropy (BCE) between rendered and ground truth masks; $\varphi$: weighting factor between RGB and segmentation tasks, set as 1 by default.

FruitNeRF adopts a joint training strategy, where a single model is trained to simultaneously predict RGB colors and segmentation masks. This is achieved through architectural modifications—specifically, by adding an additional segmentation head—and incorporating a dedicated loss function to supervise mask prediction. While FruitNeRF has demonstrated success in segmenting various fruit types across both synthetic and real-world datasets, it still faces several limitations.

The geometry in FruitNeRF is primarily inferred from the smooth gradients in RGB images, while binary segmentation masks—characterized by sharp transitions and limited texture—provide minimal supervisory signal for learning a coherent density field. As a result, the joint training process must balance two inherently different objectives, which can lead to suboptimal performance in both RGB reconstruction and mask prediction. This problem becomes more pronounced when the objects of interest are small, since the segmentation loss contributes less to the overall optimization compared to the RGB loss, slowing down or weakening the learning of accurate masks.

In addition to training limitations, the current implementation of FruitNeRF is tied to an earlier version of Nerfstudio, which may pose integration challenges with modern toolchains. This reliance on deprecated components can complicate deployment, particularly for researchers seeking to apply the method to custom datasets or real-world pipelines. While the core concept remains promising, these technical hurdles may limit its broader accessibility and adoption.

InvNeRF-Seg tries to address the problems come along with the two frameworks mentioned above, simplify the model training process by recycling the original model architecture as it is from Nerfstudio and achieve better performance in 3D segmentation at the same time.

## III. PROPOSED APPROACH

### A. Motivation

NeRF models rely heavily on smooth photometric gradients—such as color, shading, and texture—to infer scene geometry [2]. As a result, directly training NeRF using binary segmentation masks, which lack these properties, typically leads to failure. We explored modifying the RGB images based on the masks, changing the background pixels to lower value ranges while increase the value ranges of masked objects to higher ranges. Once we have the NeRF model, the segmented objects can be separated from the background through setting a threshold. While this preserves the texture information of the input images, it still fails to produce usable reconstructions when training from scratch, likely due to ambiguity in geometry estimation and insufficient gradient signals. However, a successful NeRF model was derived when fine-tune a pretrained normal RGB NeRF model on the these arbitrarily modified RGB images.

To minimizing the complexities of arbitrarily modifying the RGB image, we considered a two-stage approach to fine-tune directly on the binary mask. Since a NeRF trained on RGB images already encodes accurate geometry, we hypothesized that fine-tuning the model on 2D binary masks—without changing the model or loss functions—could shift the density field to focus on foreground objects. Surprisingly, this simple fine-tuning strategy proved highly effective, resulting in high-quality segmentation in both rendered 2D masks and 3D point clouds.

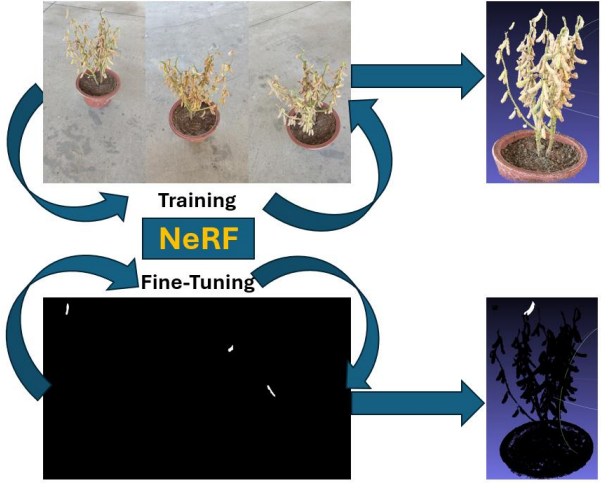

Figure 1. Graphical abstract illustrating the concept of InvNeRF-Seg.

### B. Pipeline formulation

1) **Training a standard NeRF model based on RGB images**

    NeRF learns a continuous function $f_\theta$ to map a 3D spatial coordinate $X \in R^3$ and a viewing direction $d \in R^3$ to an emitted color $c \in R^3$ and a volume density $\sigma \in R_{\geq 0}$ based on multi-view 2D RGB images with known camera poses:
    $$f_\theta(X, d) \rightarrow (c, \sigma)$$
    here, $\theta$ denotes the learnable parameters of the network.

    Points are sampled from the rays marched from each pixel to camera origin within the distance between near $a$ and far $b$ planes and then rendered to RGB values to be evaluated with input RGB ones to guide the convergence of a NeRF model:
    $$\hat{C}(r) = \int_a^b T(t)\sigma(t)c(t)dt$$
    where
    $$T(t) = \exp\left(-\int_a^t \sigma(s)ds\right)$$

    The training MSE between rendered and ground-truth RGB values is calculated as follows:
    $$l_{RGB} = \sum_{r \in R} \|\hat{C}(r) - C_{gt}(r)\|^2$$

2) **Fine-tune the trained standard NeRF using the segmented 2D masks**

    The model architecture and loss functions are unchanged compared to a standard NeRF model. The same mapping function of NeRF is shared by InvNeRF-Seg:
    $$f_\theta(X, d) \rightarrow (c, \sigma)$$
    During fine-tuning, we replace RGB images with binary masks formatted as RGB-like inputs (i.e., [1,1,1] for object, [0,0,0] for background). The loss remains as MSE:
    $$l_{mask} = \sum_r \|\hat{C}(r) - M_{gt}(r)\|^2$$
    where: $\hat{C}(r)$ rendered RGB prediction (interpreted as soft mask); $M_{gt} \in \{0,1\}^3$ is the binary mask target; $l_{mask}$: drives density to align with object regions

    No semantic supervision or binary loss (eg. BCE) is introduced in InvNeRF-Seg. The model is implicitly encouraged to reshape its density field to align with the masked object regions, since only those regions contribute to the loss. This formulation preserves the original NeRF architecture, training pipeline, and loss function, but shifts the model's learning objective via input substitution alone. Unlike previous approaches, InvNeRF-Seg does not require modifying NeRF's architecture, making it easier to integrate into existing NeRF pipelines.

## IV. EXPERIMENTS AND RESULTS

In this section, the datasets for InvNeRF-Seg and the experimental setup is first reviewed. Thereafter, comparisons of the results with other studies are provided to demonstrate the effectiveness of InvNeRF-Seg.

### A. Dataset

We used the apple and peach datasets from synthetic fruit trees in the publicly available FruitNeRF Dataset as testing examples. Details about the data creation can be found in FruitNeRF [5]. Since the dataset includes segmentation masks, camera intrinsics and poses, no additional processing was required.

In learning-based models, convergence is generally more difficult when the object class occupies a significantly smaller portion of the image compared to the background. To further challenge the models, we self-curated a soybean video dataset and designed two focused tasks: (1) binary instance-level 3D segmentation of a single soybean pod, and (2) multi-class instance-level 3D segmentation of two pods from the same plant.

Standard preprocessing was performed, including frame extraction from video using *ffmpeg* [16] at 5 frames per second and followed by camera pose estimation using *colmap* [17]. To generate pod segmentation masks, we used the AnyLabeling annotation tool with Segment Anything (MobileSAM) [18] integrated for automatic mask generation.

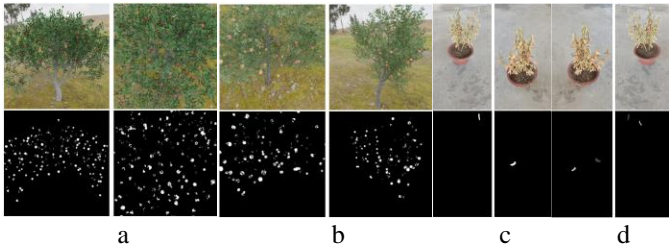

Figure 2. Examples of the synthetic data from FruitNeRF and real-world data of a soybean plant: apple (a), peach trees (b) and a Soybean plant with a single pod (c) and two pods (d) annotated in different views (top row) and segmented fruit/pod masks (bottom row).

*B. Implementation details*

The model architecture of InvNeRF-Seg remains identical to the original Nerfacto implementation from Nerfstudio. However, several hyperparameters were adjusted to ensure that a high-quality RGB NeRF model could be obtained prior to fine-tuning. Additionally, the clustering procedure was modified from that used in FruitNeRF to improve object separation and counting accuracy.

1) **NeRF in 3D segmentation**

To minimize implementation complexity while achieving strong segmentation performance, we adopted the recommended Nerfacto model from from Nerfstudio [19] and kept all configuration settings at their default values. The only exception was the far plane, which was reduced from the default of 1000 to 50 for both the apple and peach datasets to improve learning efficiency by constraining the volumetric sampling space.

For the soybean plant dataset, the default configuration was insufficient to reconstruct high-quality 3D RGB point clouds. To address this, we increased the model capacity by raising four key parameters: *hidden_dim*, *hidden_dim_color*, *hidden_dim_transient*, and *appearance_embed_dim*, from either 64 or 32 to 128. This modification enabled the model to better capture the complex and diverse visual features in the scene. Additionally, the far plane was reduced to 2 to further facilitate effective convergence on this tightly bounded, object-centric dataset.

In practice, we found that this reduction was not only beneficial for learning efficiency, but also critical for stable fine-tuning, especially when the target object occupies a small portion of the scene. For instance, fine-tuning for single-pod segmentation consistently failed when using a large far plane (e.g., 1000), but succeeded reliably when the far plane was reduced to 2. This highlights that volume bounding is essential for focusing supervision and stabilizing field refinement during segmentation of localized objects.

To ensure a fair comparison, all modifications applied to InvNeRF-Seg were also applied to FruitNeRF during training.

2) **Implementation of SA3D for comparison**

To provide a consistent comparison, we implemented SA3D by back-projecting 2D segmentation masks into 3D using accumulated field density values. Unlike the original SA3D, which may rely on volume-rendered weights, we extract the 3D points by thresholding raw field density. This choice allows for denser object reconstruction, as weights typically highlight only the object's surface, resulting in shell-like representations. In contrast, field density provides a fuller volumetric signal, capturing the interior of the target object more completely, which is particularly beneficial for object-level analysis such as clustering or counting.

3) **Clustering and counting**

After point cloud generation, the fruit points were first separated from the background to reduce computational overhead. We then followed the same clustering pipeline as described in FruitNeRF, which includes outlier removal, voxel-based downsampling, DBSCAN clustering, and merging of small clusters. At this stage, we introduced an additional procedure to split large clusters: the number of sub-clusters was determined based on the variance within each cluster and the total number of points it contained. Depending on these metrics, a cluster was either split into two or three or four sub-clusters, or left unchanged. We employed Agglomerative Clustering from scikit-learn (v1.6) [20] to reassign the points within these large clusters to their respective sub-clusters.

*C. Evaluation metrics*

Peack signal-to-noise (PSNR) [21] and Intersection over Union (IoU) [22] were used to evaluate the quality of the rendered 2D RGB images and masks, respectively, during training. In addition, the accuracy of fruit counting was used to evaluate the quality of generated 3D segmentation point cloud.

*D. Statistical analysis*

A two-tailed t-test was performed using SciPy (v1.15.2) [23] to evaluate the statistical significance of differences in PSNR and IoU scores between the compared models.

*E. Experimental environment*

All the experiments were carried out on a Windows 11 pro platform with Intel Core i9 5 GHz × 16 processor (CPU) (Intel

Corporation, Santa Clara, CA, USA), 64 GB of RAM, and a graphics processing unit (GPU) NVIDIA RTX A5500 with 16 GB of RAM (Nvidia Corporation, Santa Clara, CA, USA).

*F. Results and analysis*

1) **Evaluation on synthetic dataset**

The quality of the rendered RGB images for both apple and peach datasets was comparable between InvNeRF-Seg and FruitNeRF, as indicated by the non-significant differences in PSNR values—despite slightly higher scores from InvNeRF-Seg (Fig. 3 and Fig. 4). This outcome is expected, as both models share the same architecture for RGB-based 3D reconstruction.

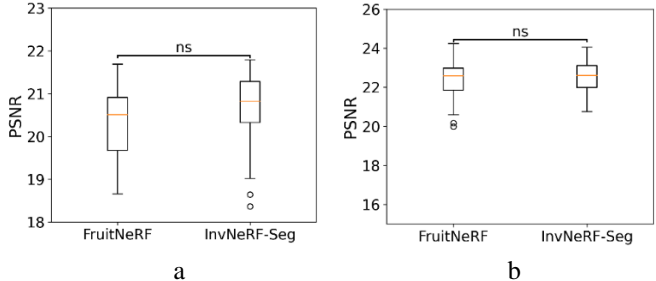

Figure 3. Comparison of PSNR between FruitNeRF and InvNeRF-Seg for rendering RGB images of apple (a) and peach (b). 'ns' indicates no statistical significance.

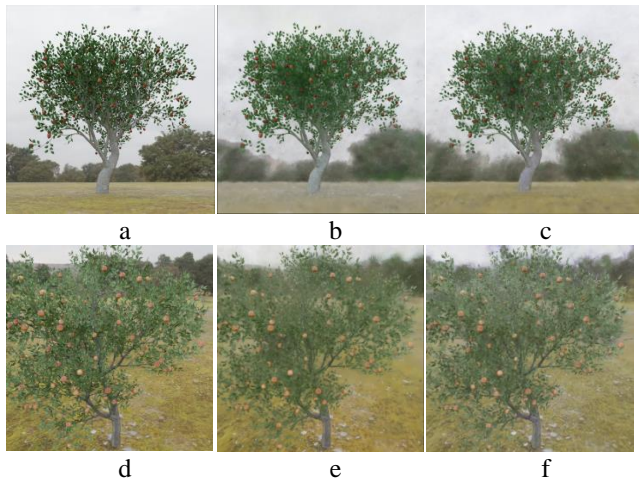

Figure 4. Examples of apple (1$^{st}$ row) and peach (2$^{nd}$ row) images rendered from FruitNeRF and InvNeRF-Seg model. Ground truth RGB images (a, d); images rendered from FruitNeRF (b, e); images rendered from InvNeRF (c, f), respectively.

On the contrary, the IoU of the rendered masks for both apple and peach datasets was significantly higher in InvNeRF-Seg compared to FruitNeRF, with *p* values below 0.001—highlighting the superior segmentation accuracy of InvNeRF-Seg (Fig. 5 and Fig. 6).

This is expected, as FruitNeRF remains dependent on RGB supervision, treating segmentation as a secondary task. Both the RGB and segmentation heads share the same feature MLP, which may limit the segmentation head's capacity to specialize. Consequently, segmentation performance may degrade to basic foreground-background separation.

In contrast, InvNeRF-Seg fine-tunes the density field directly using ground truth masks, allowing for a more precise alignment between the learned radiance field and object boundaries. This targeted refinement results in substantially cleaner and more accurate 3D segmentations.

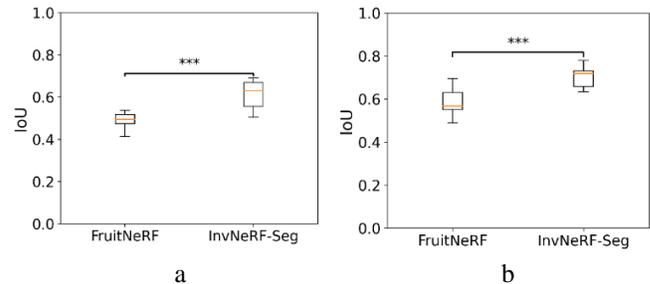

Figure 5. Comparison of IoU between FruitNeRF and InvNeRF-Seg for predicting masks of apple (a) and peach (b). '***' indicates statistical significance at p < 0.001.

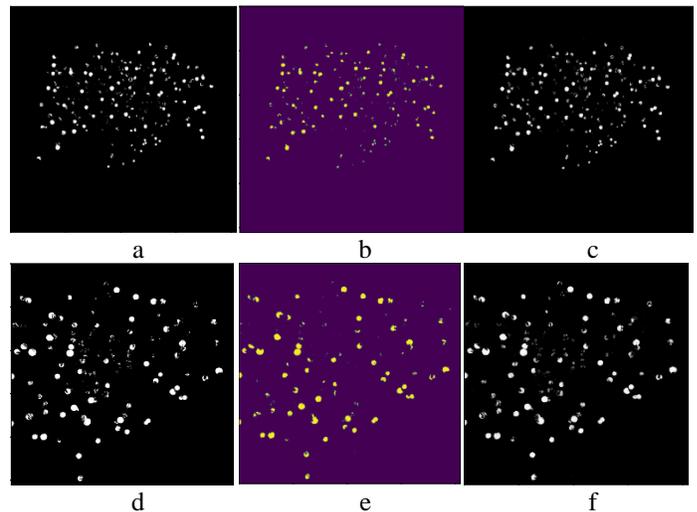

Figure 6. Examples of apple (1$^{st}$ row) and peach (2$^{nd}$ row) masks rendered from FruitNeRF and InvNeRF-Seg model. Ground truth masks (a, d); masks predicted from FruitNeRF (b, e); masks rendered from InvNeRF-Seg (c, f), respectively.

The object counting accuracy of InvNeRF-Seg was substantially higher than that of SA3D on both the apple and peach datasets. While FruitNeRF and InvNeRF-Seg performed similarly in counting apples, InvNeRF-Seg achieved higher accuracy in counting peaches (Table 1). Visualizations of the segmented and clustered 3D point clouds for all three methods are shown in Figure 7 (apple) and Figure 8 (peach). The segmented point clouds generated by InvNeRF-Seg are visibly cleaner and contain fewer artifacts, which may contribute to its improved object counting accuracy compared to both FruitNeRF and SA3D.
.

Table 1. Apple and peach counting based on segmented point cloud by three different models: FruitNeRF, InvNeRF-Seg and SA3D.

|  | FruitNeRF (pred/gt) | InvNeRF-Seg (pred/gt) | SA3D (pred/gt) |
|---|---|---|---|
| Apple | 282/283 | 282/283 | 267/283 |
| Peach | 157/152 | 152/152 | 170/152 |

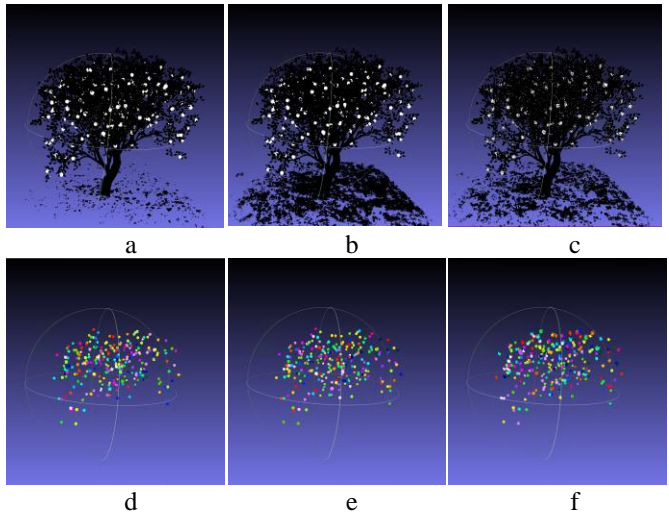

Figure 7. Visualization of segmented point cloud of apple from FruitNeRF, InvNeRF-Seg and SA3D, respectively. Generated by FruitNeRF (a, d); generated from InvNeRF-Seg (b, e); generated from SA3D (c, f), respectively.

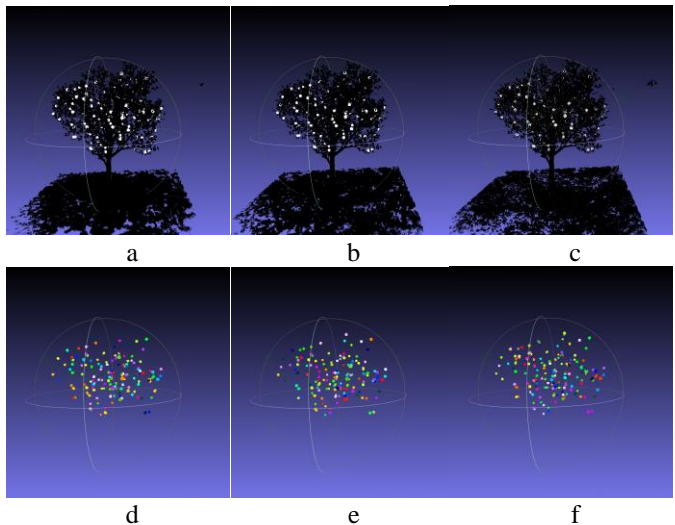

Figure 8. Visualization of segmented point cloud of peach from FruitNeRF, InvNeRF-Seg and SA3D, respectively. Generated by FruitNeRF (a, d); generated from InvNeRF-Seg (b, e); generated from SA3D (c, f), respectively.

2) **Challenging three models with binary instance-level 3D segmentation**

To evaluate the robustness of the proposed method under challenging conditions, we tested its ability to perform binary instance-level 3D segmentation by isolating a single pod from a soybean plant. InvNeRF-Seg again showed comparable performance in rendering RGB images (Fig. 9), while producing substantially higher-quality segmentation masks during training compared to FruitNeRF (Fig. 10). The segmented 3D single pod generated by InvNeRF-Seg also appeared significantly cleaner and less noisy than those produced by both FruitNeRF and SA3D, based on visual inspection (Fig. 11).

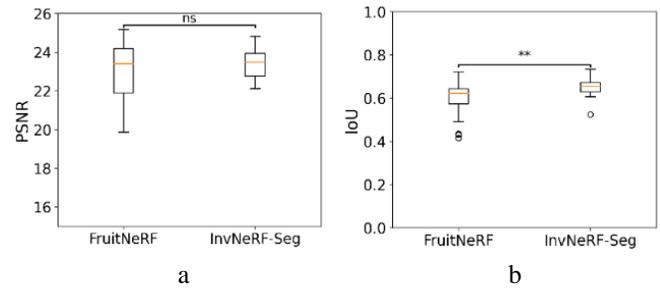

Figure 9. Comparison of PSNR (a) on rendered RGB images and IoU (b) of rendered masks for a single pod from a soybean plant between FruitNeRF and InvNeRF-Seg. 'ns' indicates no statistical significance; '**' indicates statistical significance at $p < 0.01$.

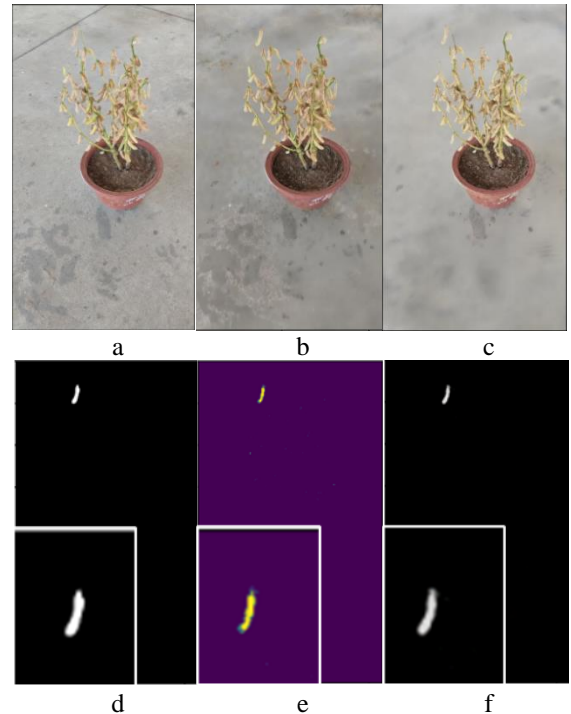

Figure 10. Example images and masks of a single pod from a soybean plant predicted by the FruitNeRF and InvNeRF-Seg models. Ground truth RGB image (a) and mask (d); RGB image (b) and mask (e) rendered by FruitNeRF; RGB image (c) and mask (f) predicted by InvNeRF-Seg, respectively.

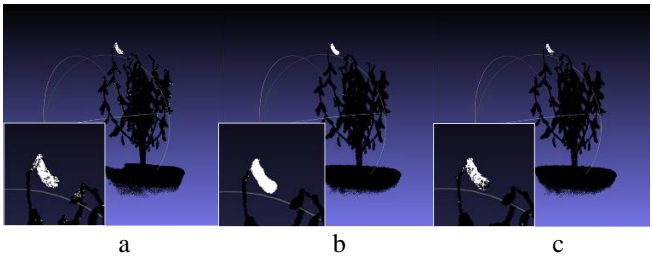

Figure 11. Segmented single-pod point clouds from a soybean plant using FruitNeRF (a), InvNeRF-Seg (b), and SA3D (c).

3) **Field Density change analysis**

The volumetric field density before and after fine-tuning was analyzed to better understand the internal behavior of InvNeRF-Seg. Targeted object regions consistently show increased density after fine-tuning, leading to sharper and more localized peaks along sampled rays (Figure 12). In contrast, non-targeted objects and background areas exhibit field density suppression or small fluctuations around zero, which results in minimal accumulated weights and effective background removal (Figure 13). These patterns confirm that InvNeRF-Seg semantically reshapes the radiance field without requiring architectural changes.

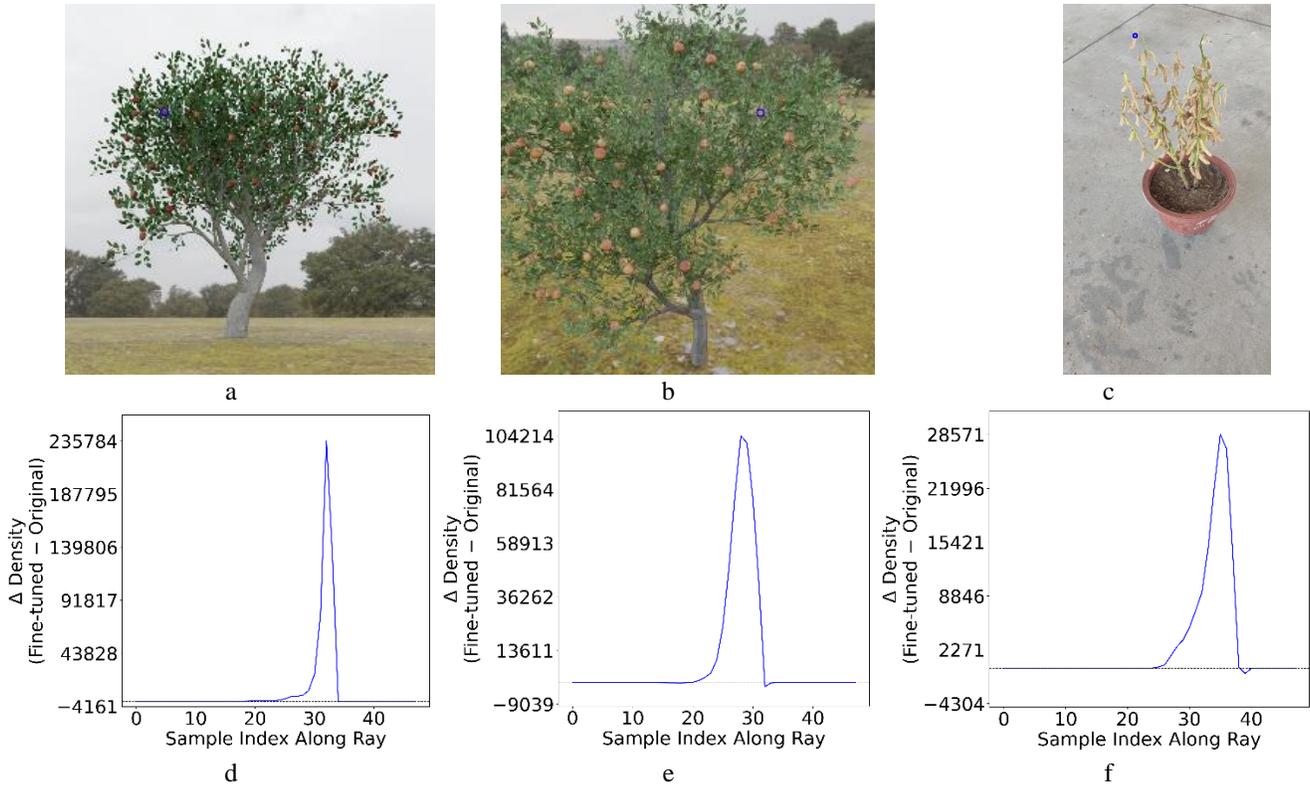

Figure 12. Field density changes in targeted object regions after fine-tuning with InvNeRF-Seg. Top row: example input pixels selected from fruit areas in apple (a), peach (b), and soybean pod (c) datasets. Bottom row: corresponding per-ray field density differences before and after fine-tuning (d–f). Positive values indicate an increase in field density, while negative values indicate suppression.

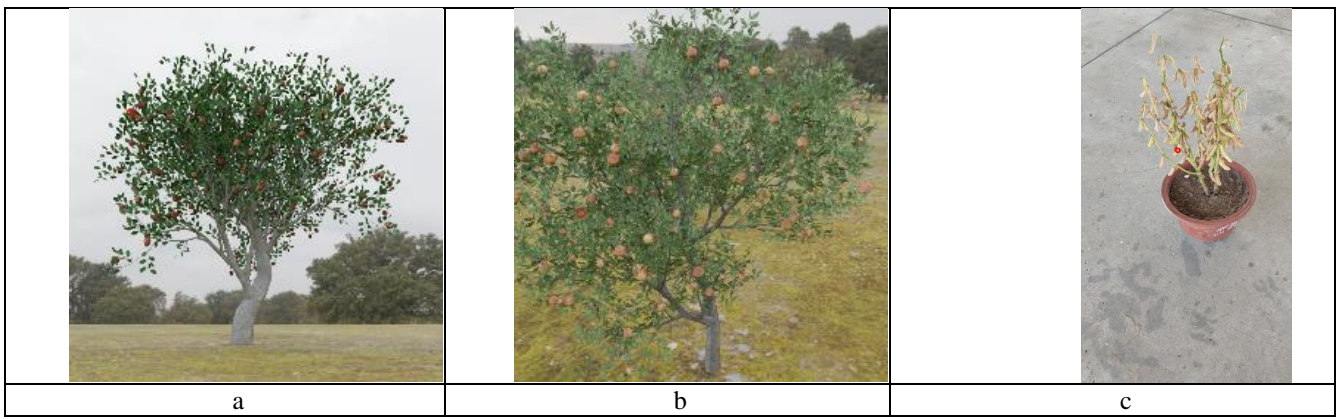

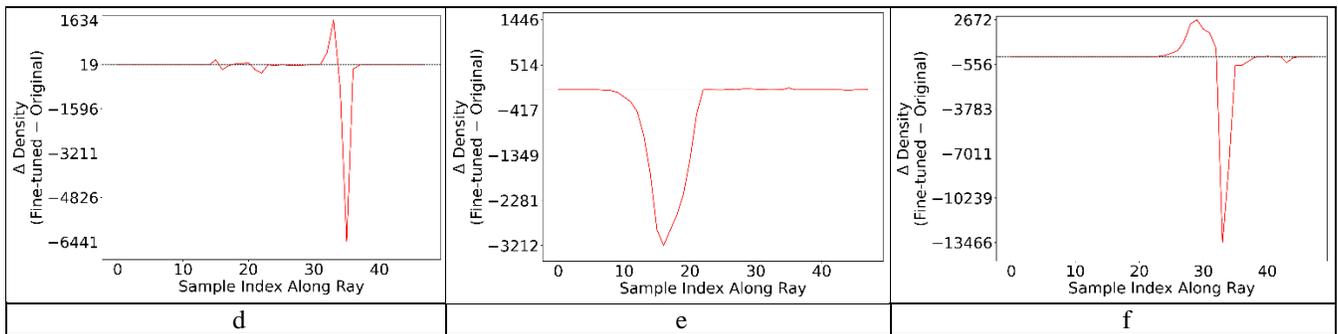

Figure 13. Field density changes in background and non-targeted object regions after fine-tuning with InvNeRF-Seg. Top row: example input pixels selected from background areas in apple (a), peach (b), and soybean pod (c) datasets. Bottom row: corresponding per-ray field density differences before and after fine-tuning (d–f). Positive values indicate an increase in field density, while negative values indicate suppression.

4) **Extension of InvNeRF-Seg to multi-class instance-level 3D segmentation**

Although InvNeRF-Seg was originally designed for binary segmentation, we further tested its ability to handle multi-class instance-level segmentation by segmenting two distinct pods from the same soybean plant.

Despite using the same MSE loss and unchanged model architecture, the method produced promising results for double-pod segmentation. However, its performance was slightly lower than in the binary case, as evidenced by some misclassified points along the instance boundaries (Fig. 14a and c). SA3D also performed reasonably well in this task, though its outputs contained notable noise across both object and background regions (Fig. 14b and d).

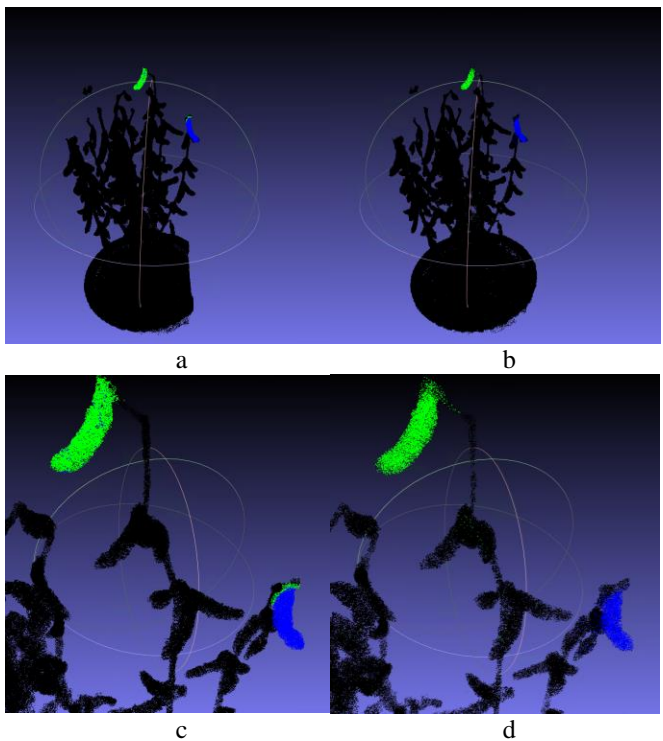

Figure 14. Segmented double-pod point clouds from a soybean plant using InvNeRF-Seg (original (a), zoomed-in (c)) and SA3D (original (b), zoomed-in (d)).

## V. DISCUSSION

Our experiments across both synthetic and real-world datasets demonstrate that InvNeRF-Seg, a two-step fine-tuning strategy that leverages a pre-trained RGB NeRF, consistently outperforms existing 3D segmentation methods—namely SA3D [3] and the joint training-based FruitNeRF [5]—in terms of segmentation accuracy, 3D point cloud quality, and object-level counting performance.

Quantitatively, InvNeRF-Seg achieves significantly higher IoU scores than FruitNeRF, particularly in scenarios where the segmentation masks represent only a small portion of the image, such as single-object cases. Qualitatively, the segmented 3D point clouds are cleaner and contain fewer artifacts than those produced by either FruitNeRF or SA3D. This improvement translates into more accurate downstream object counting. In contrast, SA3D's reliance on post-hoc density or weight filtering leads to noisier segmentations, while FruitNeRF, despite being trained end-to-end, often struggles to balance RGB reconstruction and segmentation learning.

The success of InvNeRF-Seg stems from several key factors. First, it decouples geometry learning from semantic supervision. NeRF models rely on smooth color gradients in RGB images to learn scene geometry; binary segmentation masks with hard transitions provide limited geometric cues and can hinder training if used too early. Joint learning, as in FruitNeRF, forces the model to optimize for conflicting objectives—RGB fidelity and semantic separation—using shared features, which can lead to suboptimal performance on both tasks [24]. In contrast, InvNeRF-Seg preserves geometric integrity by first optimizing for RGB reconstruction, and only then introduces semantic supervision during fine-tuning.

Second, InvNeRF-Seg benefits from more stable and efficient training. The fine-tuning stage starts from a well-initialized density field, requiring only minor refinements to align with object boundaries. As a result, convergence is faster and more robust to hyperparameter variation than joint training methods. FruitNeRF often requires careful tuning of loss weights to balance segmentation and reconstruction, whereas InvNeRF-Seg avoids this entirely.

Third, InvNeRF-Seg benefits from the ability to leverage a well-initialized radiance field during fine-tuning. This enables the model to make minimal, targeted adjustments that improve segmentation accuracy without destabilizing the learned geometry. Our field-level analysis supports this behavior: after fine-tuning, the density of object regions consistently increases, while background and non-targeted elements are suppressed—indicating a semantic restructuring of the radiance field. This effect is particularly pronounced in sparse-target scenarios, such as single-pod segmentation, where background elements like leaves and stems are confidently de-emphasized.

We also observed that volume control plays a critical role in successful fine-tuning. Specifically, setting an appropriately small far plane is crucial when the object of interest occupies only a localized region of the scene. For instance, a far plane of 1000 failed to produce meaningful results in the single-pod segmentation task, whereas reducing it to 2 enabled successful convergence. Importantly, the far plane must be consistent between RGB NeRF training and fine-tuning. A mismatch in spatial sampling regions can prevent the model from leveraging the learned geometry for semantic refinement.

Beyond binary segmentation, we evaluated InvNeRF-Seg's flexibility in multi-class instance-level segmentation by assigning different intensity values to each object in the mask. Despite using the same MSE loss and unchanged architecture, the method produced promising results in double-pod segmentation tasks. However, we observed label ambiguity near object boundaries, likely due to overlapping predicted values and the regression nature of MSE [25]. This reveals a trade-off between architectural simplicity and class separation accuracy. Incorporating class-aware losses such as cross-entropy may improve multi-class performance, although this would break the "zero-change" design. Notably, extending FruitNeRF to support multi-class segmentation would require additional architectural and loss-function modifications, further highlighting the simplicity and generalizability of our approach.

Finally, the broader implications of this work extend beyond segmentation. The results reinforce the idea that geometry learning is a critical foundation for 3D semantic understanding. InvNeRF-Seg demonstrates that object-level 3D segmentation can be achieved without architectural complexity, making it practical and accessible for a wide range of real-world applications. This zero-change, modular strategy has the potential to serve as a flexible component in future interactive 3D perception systems.

## VI. CONCLUSIONS

We proposed InvNeRF-Seg, a two-step, zero-change fine-tuning strategy for NeRF-based 3D object segmentation that preserves the original network architecture and loss function. By decoupling RGB reconstruction and segmentation learning, our method achieves superior segmentation quality and training stability compared to both the joint-learning approach (FruitNeRF) and the post-processing method (SA3D).

InvNeRF-Seg consistently produces high-quality segmented 3D point clouds and accurate object-level clustering results. Field-level analysis reveals reinforced density in target regions and suppression of background content, confirming effective semantic refinement within the radiance field. Moreover, the method extends naturally to both binary instance-level and multi-class instance-level segmentation tasks, demonstrating strong generalizability across different scene complexities and object densities.

It is important to note that this work focuses solely on the segmentation process, assuming access to accurate multi-view RGB images and corresponding 2D segmentation masks. The generation of these masks is beyond the scope of this study. However, with the increasing availability of robust 2D segmentation models, InvNeRF-Seg provides a simple, architecture-invariant bridge between 2D and 3D segmentation—lowering the barrier to high-quality 3D understanding for researchers and practitioners across domains.

REFERENCES

[1] A. Pumarola, E. Corona, G. Pons-Moll, and F. Moreno-Noguer, 'D-nerf: Neural radiance fields for dynamic scenes', in *Proceedings of the IEEE/CVF conference on computer vision and pattern recognition*, 2021, pp. 10318–10327.

[2] B. Mildenhall, P. P. Srinivasan, M. Tancik, J. T. Barron, R. Ramamoorthi, and R. Ng, 'Nerf: Representing scenes as neural radiance fields for view synthesis', *Commun ACM*, vol. 65, no. 1, pp. 99–106, 2021.

[3] J. Cen *et al.*, 'Segment anything in 3d with radiance fields', *arXiv preprint arXiv:2304.12308*, 2023.

[4] S. Zhi, T. Laidlow, S. Leutenegger, and A. J. Davison, 'In-place scene labelling and understanding with implicit scene representation', in *Proceedings of the IEEE/CVF International Conference on Computer Vision*, 2021, pp. 15838–15847.

[5] L. Meyer, A. Gilson, U. Schmid, and M. Stamminger, 'FruitNeRF: A Unified Neural Radiance Field based Fruit Counting Framework', in *2024 IEEE/RSJ International Conference on Intelligent Robots and Systems (IROS)*, IEEE, 2024, pp. 1–8.

[6] Y. Jiang, G. Liu, Z. Huang, B. Yang, and W. Yang, 'Geometry perception and motion planning in robotic assembly based on semantic segmentation and point clouds reconstruction', *Eng Appl Artif Intell*, vol. 130, p. 107678, 2024.

[7] Y. Wen *et al.*, 'AR-Light: Enabling Fast and Lightweight Multi-user Augmented Reality via Semantic Segmentation and Collaborative View Synchronization', *IEEE Transactions on Computers*, 2025.

[8] Y. Zhuang *et al.*, '3D-SeqMOS: A Novel Sequential 3D Moving Object Segmentation in Autonomous Driving', *IEEE Transactions on Intelligent Transportation Systems*, 2024.

[9] W. Li *et al.*, 'Fairdiff: Fair segmentation with point-image diffusion', in *International Conference on Medical Image Computing and Computer-Assisted Intervention*, Springer, 2024, pp. 617–628.

[10] R. Yang *et al.*, '3D-based precise evaluation pipeline for maize ear rot using multi-view stereo reconstruction and point cloud semantic segmentation', *Comput Electron Agric*, vol. 216, p. 108512, 2024.

[11] C. R. Qi, H. Su, K. Mo, and L. J. Guibas, 'Pointnet: Deep learning on point sets for 3d classification and segmentation', in *Proceedings of the IEEE conference on computer vision and pattern recognition*, 2017, pp. 652–660.

[12] C. R. Qi, L. Yi, H. Su, and L. J. Guibas, 'Pointnet++: Deep hierarchical feature learning on point sets in a metric space', *Adv Neural Inf Process Syst*, vol. 30, 2017.

[13] T. Miao, W. Wen, Y. Li, S. Wu, C. Zhu, and X. Guo, 'Label3DMaize: toolkit for 3D point cloud data annotation of maize shoots', *Gigascience*, vol. 10, no. 5, p. giab031, 2021.

[14] L. Pan, 'ECG: Edge-aware point cloud completion with graph convolution', *IEEE Robot Autom Lett*, vol. 5, no. 3, pp. 4392–4398, 2020.

[15] C. Liu *et al.*, 'Context-aware network for semantic segmentation toward large-scale point clouds in urban environments', *IEEE Transactions on Geoscience and Remote Sensing*, vol. 60, pp. 1–15, 2022.

[16] S. Tomar, 'Converting video formats with FFmpeg', *Linux journal*, vol. 2006, no. 146, p. 10, 2006.

[17] J. L. Schonberger and J.-M. Frahm, 'Structure-from-motion revisited', in *Proceedings of the IEEE conference on computer vision and pattern recognition*, 2016, pp. 4104–4113.

[18] C. Zhang *et al.*, 'Faster segment anything: Towards lightweight sam for mobile applications', *arXiv preprint arXiv:2306.14289*, 2023.

[19] M. Tancik *et al.*, 'Nerfstudio: A modular framework for neural radiance field development', in *ACM SIGGRAPH 2023 conference proceedings*, 2023, pp. 1–12.

[20] O. Kramer and O. Kramer, 'Scikit-learn', *Machine learning for evolution strategies*, pp. 45–53, 2016.

[21] J. Korhonen and J. You, 'Peak signal-to-noise ratio revisited: Is simple beautiful?', in *2012 Fourth international workshop on quality of multimedia experience*, IEEE, 2012, pp. 37–38.

[22] R. Girshick, 'Fast r-cnn', in *Proceedings of the IEEE international conference on computer vision*, 2015, pp. 1440–1448.

[23] P. Virtanen *et al.*, 'SciPy 1.0: fundamental algorithms for scientific computing in Python', *Nat Methods*, vol. 17, no. 3, pp. 261–272, 2020.

[24] T. Standley, A. Zamir, D. Chen, L. Guibas, J. Malik, and S. Savarese, 'Which tasks should be learned together in multi-task learning?', in *International conference on machine learning*, PMLR, 2020, pp. 9120–9132.

[25] J. Zhao *et al.*, 'Deep-learning-based multispectral image reconstruction from single natural color RGB image—Enhancing UAV-based phenotyping', *Remote Sens (Basel)*, vol. 14, no. 5, p. 1272, 2022.